# Open Brain AI. Automatic Language Assessment

**Charalambos Themistocleous**
Department of Special Needs Education, University of Oslo
Helga Engs hus 4.etg, Sem Sælands vei 7 0371 OSLO
charalampos.themistokleous@isp.uio.no

**Abstract**

Language assessment plays a crucial role in diagnosing and treating individuals with speech, language, and communication disorders caused by neurogenic conditions, whether developmental or acquired. However, current assessment methods are manual, laborious, and time-consuming to administer and score, causing additional patient stress. To address these challenges, we developed Open Brain AI (https://openbrainai.com). This computational platform harnesses innovative AI techniques, namely machine learning, natural language processing, large language models, and automatic speech-to-text transcription, to automatically analyze multilingual spoken and written speech productions. This paper discusses the development of Open Brain AI, the AI language processing modules, and the linguistic measurements of discourse macro-structure and micro-structure. The fast and automatic analysis of language alleviates the burden on clinicians, enabling them to streamline their workflow and allocate more time and resources to direct patient care. Open Brain AI is freely accessible, empowering clinicians to conduct critical data analyses and give more attention and resources to other critical aspects of therapy and treatment.

## 1. Introduction

Speech, language, and communication disorders affect both children and adults. In a year, almost 7.7% (one in twelve) of US children ages 3-17 were diagnosed with speech and language-related disorders (Law, Boyle, Harris, Harkness, & Nye, 2000). Post-stroke aphasia appears in 21–38% of acute stroke patients (Berthier, 2005; Pedersen, Vinter, & Olsen, 2004). Impaired speech, language, and communication can be a symptom of severe conditions, such as Alzheimer's Disease, brain tumors, stroke, and neurogenic developmental conditions (Ahmed, Haigh, de Jager, & Garrard, 2013; Meilan, Martinez-Sanchez, Carro, Carcavilla, & Ivanova, 2018; Mueller, Hermann, Mecollari, & Turkstra, 2018; Petersen et al., 1999; Ribeiro, Guerreiro, & de Mendonça, 2007; Themistocleous, Eckerström, & Kokkinakis, 2020; Themistocleous & Kokkinakis, 2019; Weiss et al., 2012). Speech, language, and communication disorders challenge individuals' ability to express themselves effectively and participate in social interactions, leading to social isolation, depression, and inferior quality of life. Therefore, early screening and assessing individuals for speech, language, and communication disorders is crucial for effective diagnosis, prognosis, and treatment efficacy assessment (Strauss, Sherman, Spreen, & Spreen, 2006, pp. 891-962). Also, language assessment can supplement the Assessment of cognitive domains, such as memory and attention, and provide measures correlating with these cognitive domains (Battista et al., 2017; Cohen & Dehaene, 1998; Lezak, 1995; Thomas, Billon, & Hazif-Thomas, 2018).

Assessing speech, language, and communication informs about the neurolinguistic functioning of individuals as well as their cognitive condition. It can account for pathology in cortical areas associated with written speech production and comprehension and thus inform treatment approaches (de Aguiar et al., 2020; Fischer-Baum & Rapp, 2014; K. Neophytou, Wiley, Rapp, & Tsapkini, 2019; Purcell & Rapp, 2018; Rapp & Fischer-Baum, 2015; Themistocleous, Neophytou, Rapp, & Tsapkini, 2020; Tsapkini et al., 2018). Therefore, speech, language, and communication assessments have always been the bedrock of neurocognitive and neurolinguistic assessments for patients.

Nevertheless, manual assessments and tests commonly used in the clinic, such as the Boston Naming Test (BNT; Kaplan, Goodglass, & Weintraub, 2001), Western Aphasia Battery-Revised (WAB-R; Kertesz (2006)), Boston Diagnostic Aphasia Examination (BDAE; Goodglass & Kaplan, 1983), Psycholinguistic Assessment of Language Processing in Aphasia (PALPA; Kay, Lesser, & Coltheart, 1992) can be cumbersome and time-consuming to deliver and score and can stress the patients or students who take the assessments.

Importantly, manual assessments are usually single-domain assessments. For example, they assess sentence repetition, fluency, and confrontational naming for nouns, which is not an ecological depiction of language communication in terms of the setting they are produced and information complexity. Also, they are conducted in controlled environments, which do not reflect real-life situations and have a narrow scope, focusing on specific linguistic domains. For example, naming tasks were not developed to provide information about syntactic production, and picture description tasks, such as the Cookie-Theft Picture Description task from the Boston Diagnostic Aphasia Examination (Goodglass & Kaplan, 1983; Goodglass, Kaplan, & Barresi, 2001), provide only limited information on how individuals produce different sentence types, such as questions and commands, morphological categories, such as verbs in the past or future, and deontic modality structure expressing wish and volition.

They also focus on written speech, whereas spoken production is evaluated only from the point of view of the clinician's perception. However, clinicians' perceptions may miss differences from the largest population. It does not constitute a consistent measure of the phonetics of patients, as clinical judgments may vary from one timepoint to another and between clinicians. Consequently, there is a critical need for a better approach to manual assessment.

An alternative way is to employ computational tools for analyzing speech, language, and communication in naturalistic settings, such as discourse and conversation. Discourse and conversation can offer an ecological depiction of speech, language, and communication (Stark, Bryant, Themistocleous, den Ouden, & Roberts, 2022; Stark et al., 2020). For example, discourse tasks offer the opportunity to elicit multidomain linguistic data, such as measures for sentence-level discourse microstructure (e.g., morphology, syntax, semantics) and microstructure (e.g., cohesion and coherence information structure, planning, topics) (Themistocleous, 2022). Also, discourse and communication analysis can identify the effects of dementia on language (Themistocleous, 2022) and quantify language function and the impact of dementia on the cognitive representations of grammar (rules and principles) that enable speakers to produce grammatically correct sentences (Chomsky, 1965); communicative competence, which is the ability of individuals to employ language appropriately in social environments and settings, emotions, empathy, and theory-of-mind (Murray, Timberlake, & Eberle, 2007); and talk-in-interaction to identify how individuals with dementia follow the turn-taking dynamics and conventions in conversations (Sacks, Schegloff, & Jefferson, 1974; Schegloff, 1998; Schegloff, Jefferson, & Sacks, 1977).

Assessing speech, language, and communication disorders requires accurate and reliable measurements of various linguistic and acoustic parameters. In recent years, advancements in technology, particularly in Artificial Intelligence (AI), Natural Language Processing (NLP), Machine Learning (ML), acoustic analysis, and statistical modeling, have revolutionized the way clinicians and researchers evaluate and diagnose speech, language, and communication disorders. Open Brain AI utilizes AI technologies to provide practical assessment tools for speech, language, and communication disorders. Artificial Intelligence (AI) is a cover term that includes Machine Learning (ML) technologies, such as deep neural networks used for tasks such as learning patterns from data and making predictions on novel inputs, Natural Language Processing (NLP) that provides algorithms to analyze and interpret linguistic patterns, acoustic analysis, and signal processing to analyze speech recordings. AI-based systems automate tasks, such as speech transcription, language comprehension assessment, and language generation, providing clinicians with valuable tools to enhance the accuracy and efficiency of estimates.

## 2. Open Brain AI

Open Brain AI (http://openbrainai.com) is an online platform that employs computer technology and Artificial Intelligence (AI) tools for assessing speech, language, and communication. Open Brain AI analyzes spoken and written language and provides informative linguistic measures of discourse and conversation. This analysis is meant to support clinicians and speech and language therapists to assess the language functioning of their patients and offer diagnosis, prognosis, therapy efficacy evaluation, and treatment planning. Finally, Open Brain AI is a research environment for computational language assessment (CLA). It allows researchers and clinicians to collaborate, share ideas, and evaluate novel technologies for patient care and student learning.

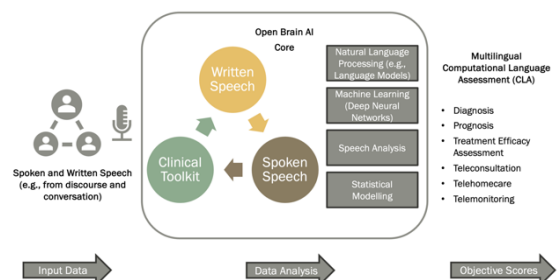

**Figure 1.** The primary components of Open Brain AI for multilingual Computational Language Assessment and Screening in a three-stage process: 1) input data, 2) data analysis using trained ML models, and 3) output objective scores.

Open Brain AI combines different computational pipelines (see Figure 1): speech-to-text, large language models, morphological taggers/parsers of the analysis of grammar, semantic analysis tools, IPA transcription tools, and acoustic analysis tools. Open Brain AI enables end-to-end spoken and written speech production analysis by combining the different computational pipelines to provide automated and objective linguistic measures. The computational pipelines of Open Brain AI resulted from our previous work and were published in other papers (Themistocleous, 2017a, 2019; Themistocleous, Eckerström, & Kokkinakis, 2018; Themistocleous, Ficek, et al., 2021; Themistocleous & Kokkinakis, 2018; Themistocleous, Neophytou, et al., 2020;

Themistocleous, Webster, Afthinos, & Tsapkini, 2020; Themistocleous, Webster, & Tsapkini, 2021). The studies will be discussed in the following section. However, it should be emphasized that this work has been under development for many years, starting from 2007. The platform relies on our ongoing research; thus, it will change over time in terms of existing tools and adding new tools, features, and components following our current study at each time point and meeting the needs. The following discusses the primary domains of analysis in Open Brain AI.

## 2.1 Language assessment

The written language assessment module processes written speech transcripts and comprehensively analyzes speech, language, and communication. It comprises two three pipelines. The first analyzes the text and elicits linguistic measures, and the second pipeline combines the linguistic measures and the text and uses them to provide discourse analysis with text recommendations. The third pipeline allows the transcription of recordings and then uses the transcripts to conduct linguistic measures and analyze them for discourse. Appendix 1 provides an example of the output of the written language assessment module with the current elicited linguistic measures.

### 2.1.1 Computational Discourse Analysis with Large Language Models

Discourse provides multidomain data on language production, perception, planning, and cognition (Cunningham & Haley, 2020; Fyndanis et al., 2018; Stark et al., 2022; Stark et al., 2020). Open Brain AI's discourse module employs large AI language Models, like GPT3. It analyzes language productions by combining the text produced by a patient and metrics from discourse, semantics, syntax, morphology, phonology, and lexical distribution elicited using NLP and machine learning. Subsequently, it combines its internal knowledge of the world based on its training to provide a comprehensive analysis of speech, language, and communication for the textual transcripts based on quantified measures from part of speech analysis, syntactic phrase identification, semantic analysis (e.g., named entity recognition), and lexical distribution.

- Computational Discourse Analysis - Macrostructure (e.g., cohesion and coherence)
- Computational Discourse Analysis - Microstructure
- Error Analysis
- Recommendations on whether there is evidence for a possible speech, language, and communication impairment.

Currently, we provide analysis for English, Danish, Dutch, Finnish, French, German, Greek, Italian, Norwegian, Portuguese, Spanish, and Swedish. Assessing written speech from discourse involves evaluating an individual's written language skills and ability to organize and convey information coherently in written form.

### 2.1.2 Linguistic Measures: Phonology, Morphology, Syntax, Semantics, and Lexicon

The first part of the output is the AI assessment discussed in the previous section. The second part of the analysis is the objective measures of written speech production that clinicians, teachers, and researchers compare a patient with a targeted population concerning discourse, phonology, morphology, syntax, semantics, and lexicon (Badecker, Hillis, & Caramazza, 1990; Breining et al., 2015; A. E. Hillis & Caramazza, 1989; Argye E. Hillis, Rapp, Romani, & Caramazza, 1990; Miceli, Capasso, & Caramazza, 1994; Stockbridge et al., 2021; Themistocleous, Ficek, et al., 2021; Tsapkini, Frangakis, Gomez, Davis, & Hillis, 2014).

Specifically, this module analyzes the text or the transcripts from the speech-to-text module and conducts measures on the following linguistic domains:

- Phonology: It elicits measures, such as the number and type of syllables and the ratio of syllables per word.
- Morphology: It provides counts and their ratio of parts of speech (e.g., verbs, nouns, adjectives, adverbs, and conjunctions) concerning the total number of words.
- Syntax: It provides counts and their ratio of syntactic constituents (e.g., noun phrases and verb phrases).
- Lexical Measures: it provides measures such as the number of words, hapax legomena, and Type Token Ratio (TTR) measures.
- Semantic Measures: It provides counts and their ratio of semantic entities in the text (e.g., persons, dates, and locations).
- Readability Measures: It provides readability measures about the text and grammar.

In our previous research, we employed morphological and syntactic evaluation to analyze transcripts using natural language processing (NLP) and to provide automated part-of-speech (POS) tagging and syntactic parsing. For example, Themistocleous, Webster, et al. (2020) analyzed connected speech productions from 52 individuals with PPA using a morphological tagger. They showed differences in POS production in patients with nfvPPA, lvPPA, and svPPA. This NLP algorithm automatically provides the part of speech category for all words individuals produce (Bird, Klein, & Loper, 2009). From the tagged corpus, they measured both content words (e.g., nouns, verbs, adjectives, adverbs) and function words (conjunctions, e.g., and, or, and but; prepositions, e.g., in, and of; determiners, e.g., the a/an, both; pronouns, e.g., he/she/it and wh-pronouns, e.g., what, who, whom; modal verbs, e.g., can, should, will; possessive ending (' s), adverbial particles, e.g., about, off, up; infinitival to, as in to do). Themistocleous, Webster, et al. (2020) showed that the POS patterns of individuals with PPA were both expected and unexpected. It showed that individuals with nfvPPA produced more content words than function words (see top left for the content words and top right for the function words). Individuals with nfvPPA made fewer grammatical words than individuals with lvPPA and svPPA. These studies demonstrate that computational tools study speech and language. Thus, they form the basis for developing assessment tools for scoring patients' language and computation performance from discourse and conversation.

## 2.2 Spoken language Analysis

The spoken language analysis module includes speech-to-text, then automatically analyzes transcribed texts concerning the different linguistic levels.

**Transcription:** Open Brain AI offers automatic transcription using an Automatic Speech Recognition (ASR) system to process audio files. The process begins by uploading an audio file on Open Brain AI. Concerning the background elements (such as hm), the platform allows two strategies to keep and consider them in the analysis: the preselected option or to remove them and automatically analyze the text transcript for grammar without them.

*Speakers Segmentation.* The Open Brain AI platform offers the option for splitting the the audio, which enables the splitting patients from clinicians in the audio recordings. When there is more than one speaker in the audio file. The diarization output is exported as a coma delimited file or Praat TextGrid for researchers wanting to perform acoustic analysis.

*Word Alignment.* The platform enables the alignment of words with the sound wave to allow further acoustic analysis for measures, such as word duration, and the elicitation of the specific acoustic measures on acoustic production. The automatically segmented sounds are exported in various formats, such as Praat TextGrids.

**Linguistic Analysis & AI Discourse Analysis.** The transcripts are further analyzed using the automatic morphosyntactic analysis and by a GPT3 Large Language Model. The subsequent analysis provides the following information:

- The module combines the text and metrics from discourse, semantics, syntax, morphology, phonology, and lexical distribution.
- The module then combines its internal knowledge of the world based on training to provide a comprehensive analysis of speech, language, and communication for the textual transcripts.
- The module analyzes discourse in several languages: English, Danish, Dutch, Finnis, French, German, Greek, Italian, Norwegian, Portuguese, Spanish, and Swedish.

**Acoustic Analysis.** Speakers pronounce sounds differently depending on age, gender, and social variety (e.g., dialect, sociolect) (Themistocleous, 2016, 2017a, 2017b, 2017c, 2019; Themistocleous, Savva, & Aristodemou, 2016). The acoustic analysis of vowels and consonants can indicate pathological speech, characterizing many patients with aphasia, especially those with apraxia of speech and other acquired and developmental speech, language, and communication disorders (Themistocleous, Eckerström, et al., 2020; Themistocleous, Ficek, et al., 2021; Themistocleous, Webster, et al., 2021). Also, variations in the production of prosody (e.g., *fundamental frequency (F0) and pauses*) indicate abnormalities in pitch control, vocal fold functioning, or neurological impairments (Themistocleous, Eckerström, et al., 2020; Themistocleous, Ficek, et al., 2021; Themistocleous, Webster, et al., 2021). The spoken speech assessment module provides transcription and grammatical analysis of these transcripts. The grammatical study replicates that of written speech productions. Namely, it offers total phonology, morphology, syntax, semantics, and lexicon scores. It provides tools that allow clinicians and researchers to assess the importance of spoken speech for patients with speech, language, and communication disorders, highlighting the unique characteristics of spoken language production and its acoustic properties and making connections to the underlying biological processes involved. Spoken speech possesses distinct characteristics that set it apart from written language. It involves the real-time production of sounds and the coordination of various physiological systems. Finally, computational tools provide a comprehensive analysis of morphology in patients with different variants of Primary Progressive Aphasia (Themistocleous, Webster, et al., 2020) and argue that computational tools could analyze naturalistic speech from discourse (Themistocleous, 2022). Computational models elicit measures from speech acoustics, spelling, morphology, syntax, and semantics.

## 2.3 The Clinical Toolkit

The clinical toolkit provides scoring tools and comprises four primary tools: i. *The semantics distance tool* relies on word embeddings to automatically score verb and noun naming tests; ii. *the phonological distance tool* facilitates the scoring of phonological errors; and the iii. *the spelling scoring tool* allows the scoring of words and non-words (Themistocleous, Neophytou, et al., 2020).

### 2.3.1 Automatic conversion to the International Phonetic Alphabet

The tool converts words written in standard orthography into the International Phonetic Alphabet. The tool provides this service in several languages, including English (US), English (UK), Arabic, Chinese, Danish, Dutch, Finnish, French, German, Greek, Hindi, Icelandic, Italian, Japanese, Korean, Norwegian, Portuguese, Russian, Spanish, and Swedish.

### 2.3.2 Spelling Scoring App

The evaluation of spelling is a complex, challenging, and time-consuming process. It relies on comparing letter-to-letter, the words spelled by the patients to the target words. The tool offers multilingual spelling assessment in several languages, including English (US), English (UK), Arabic, Chinese, Danish, Dutch, Finnish, French, German, Greek, Hindi, Icelandic, Italian, Japanese, Korean, Norwegian, Portuguese, Russian, Spanish, and Swedish. It processes both words and non-words (Themistocleous, Neophytou, et al., 2020). Specifically, Themistocleous, Neophytou, et al. (2020) developed a spelling distance algorithm that automatically compares the inversions, insertions, deletions, and transpositions required to make the target word and the response the same (Kyriaki Neophytou, Themistocleous, Wiley, Tsapkini, & Rapp, 2018; Themistocleous, Neophytou, et al., 2020). To determine phonological errors in patients with aphasia, we have developed a phonological distance algorithm that quantifies phonological errors automatically.

### 2.3.3 Phonological Scoring Tool

The tool offers multilingual phonological Assessment in several languages, including English (US), English (UK), Arabic, Chinese, Danish, Dutch, Finnish, French, German, Greek, Hindi, Icelandic, Italian, Japanese, Korean, Norwegian, Portuguese, Russian, Spanish, and Swedish. It processes both words and non-words.

### 2.3.4 Semantics Scoring Tool

The tool offers semantic distance scoring using word embeddings. Word and sentence embeddings and generative large language models based on transformer architectures predict language brain's language functioning in healthy individuals, as in studies by Fedorenko and her group (Hosseini et al., 2022). These measures cater to multiple languages, including English, Greek, Italian, Norwegian, and Swedish. Lastly, Open Brain AI offers partial support for additional languages such as Arabic, Chinese, English (UK), Hindi, Icelandic, Japanese, and Korean.

## 2.4 Multilingual Support

Open Brain AI provides multilingual support in different languages and language varieties (e.g., dialects). It offers automatic transcription and comprehensive grammar analysis in English, Norwegian, Swedish, Greek, and Italian. The complete grammar analysis extends to languages such as Danish, Dutch, Finnish, French, German, Portuguese, and Spanish. Additional languages and language varieties will be supported over time as models from the different varieties are incorporated into the platform. The ability of Open Brain AI to scale concerning new languages and language variety support highlights a critical difference between computational models over traditional manual assessment techniques. Unlike manual assessments, their translation to a new language variety will require expert knowledge for translation, standardization, and evaluation while maintaining crosslinguistic psychometric properties, such as the reliability and validity of tests. The *Open Brain AI* platform offers access to these trained models for clinicians and teachers and makes them available.

## 2.5 Diagnosis and Prognosis

An accurate diagnosis and prognosis are crucial for developing tailored intervention plans to improve their quality of life (Grasemann, Peñaloza, Dekhtyar, Miikkulainen, & Kiran, 2021; Johnson, Ross, & Kiran, 2019). Prognosing individuals with speech, language, and communication disorders involves predicting their condition's course and potential outcomes (Diogo, Ferreira, Prata, & Alzheimer's Disease Neuroimaging, 2022). The role of Open Brain AI is to assist experienced clinicians in making prognostic judgments based on their clinical expertise and knowledge of empirical research findings. For example, in our previous research, we employed machine learning models and information from acoustic production to provide a classification of patients with MCI from healthy controls from speech sounds (Themistocleous et al., 2018; Themistocleous, Eckerström, et al., 2020; Themistocleous & Kokkinakis, 2019). We have also employed measures elicited using natural language processing, namely the morphosyntactic analysis of sentences from patients (e.g., measures of parts of speech and lexical distribution) and acoustic analysis (e.g., F0, duration, pauses) to subtype patients with the PPA into their corresponding variants (Themistocleous, Ficek, et al., 2020; Themistocleous, Webster, et al., 2020).

### 2.6 Data Safety

Open Brain AI does not collect data provided for analysis. Data are analyzed on the server or locally on the user's machine. Data uploaded on the server for analysis are removed immediately after processing. Information provided in Open Brain AI for accessing the site is not shared with third parties. Open Brain AI takes data privacy and security very seriously and follows industry standards to protect the confidentiality and security of personal health information. However, no data transmission over the internet is guaranteed to be completely secure. Therefore, Open Brain AI cannot guarantee the security of any information transmitted through the service, and you use the service at your own risk. Open Brain AI provided for healthcare purposes is not intended to replace or substitute for professional medical advice, diagnosis, or treatment.

### 2.7 Discussion

By leveraging AI tools and providing multilingual assessments, Open Brain AI enables the computational analysis of written and spoken speech from discourse. So, it holds significant potential for enhancing the evaluation and treatment of patients with speech, language, and communication disorders. Clinicians gain valuable insights into an individual's cognitive and linguistic abilities, elicit objective and quantitative scores of the language domains (e.g., morphology, syntax, semantics, and lexicon), facilitate functional communication treatment, and improve therapeutic interventions. Also, tools in Open Brain AI help clinicians in everyday clinical tasks, such as scoring neurolinguistic tests.

Open Brain AI stays at the forefront of computational technology and implements recent technologies. Continued advancements in AI will further enhance our understanding of speech and language pathology and enable more effective interventions for individuals with speech, language, and communication disorders. Also, Open Brain AI promotes interdisciplinary collaboration between speech-language pathologists, neurologists, psychologists, and researchers by providing an environment allowing them to evaluate novel technologies. A multidisciplinary approach allows a rounded understanding of the underlying factors contributing to speech, language, and communication disorders. This leads to more accurate prognostic and diagnostic judgments and tailored intervention plans.

- *Language Models and Automatic NLP Analysis in the clinic.* These models allow the analysis of texts and offer two types of information. A broad description of discourse that provides an overview to the clinician of the situation. In other words, it informs the clinician about what is happening in a specific text by using the text as information and the output of the NLP analysis. This part is informative, but the analysis is not quantified. The automatic analysis also provides quantified measures of linguistic domains (Beltrami et al., 2018; Fraser et al., 2019). Therefore, *Open Brain AI* written language analysis effectively enables clinicians and researchers to evaluate a patient's ability to engage in complex linguistic tasks, such as generating ideas, organizing thoughts, and conveying them logically through writing. It provides a window into the individual's higher language functions, such as syntactic complexity, vocabulary usage, and discourse coherence. Also, the insights gained from assessing written speech guide written speech intervention planning and goal setting. By identifying specific areas of difficulty, clinicians design targeted interventions that address the patient's needs, facilitate progress, and enhance overall communication abilities.
- *Multilingual Consistency.* The accuracy of tools depends on the availability of data, which depends on language variety, to language variety. This critical problem is currently evidenced in many NLP applications, including large language models and translation systems. As such, this creates a problem with getting the same outputs for all these language varieties, so a tool employed for diagnosis is performing the same across languages. Over time this will become less of a problem as more data are becoming available and algorithms that collect and preprocess this time are becoming better with uncommon languages and language varieties.
- Acc*uracy and Effectiveness:* While the accuracy and effectiveness of the models are essential for diagnosis, such as identifying patients from non-patients or subtyping patients into groups, providing prognosis, and evaluating treatment efficacy, there is also a growing need for models that offer insights into human behavior. For instance, research has demonstrated that the fundamental frequency corresponds to intonation, while the first and second formant frequencies correspond to properties of vowel quality (Themistocleous, 2017a, 2017c). The development of classification models emphasizes the accuracy of the output, e.g., for categorizing an individual as a patient or a healthy individual, without offering a clear explanation for their decision-making process. Clinicians require models explaining why a particular classification was made, shedding light on the underlying factors influencing the decision. This interpretability empowers clinicians better understand the model's outputs and enable them to make informed treatment decisions. Open Brain AI provides models and measures that provide accurate results and interpretability. It provides both models that are accurate in terms of model performance but also provides models and scores that clinicians can employ to understand the condition of their patients.

- *Web application vs. offline analysis:* Open Brain AI facilitates research on speech and language, allowing researchers to automate their everyday workflow, e.g., working with data with a limited number of patients (McCleery, Laverty, & Quinn, 2021). It is challenging to employ a web application to automate the analysis of multiple data from different speakers or speech productions, which requires custom scripts. To address this, we have implemented offline pipelines that allow flexibility and bigger offline models to analyze complex data for researchers. Offline analysis allows us to use and train models that cannot be conducted on a server due to the high costs of loading current server infrastructures with data and large computational models.
- As such, Open Brain AI provides technologies that can support i. telehealth and teleconsultation by providing feedback to health clinicians from patients at a distance to create a better picture of a patient's condition (McCleery et al., 2021); ii. telehomecare by aiding personnel responsible for patient care about a patient's linguistic abilities, and iii. telemonitoring by providing data over time from language, and as such, it can work together with other monitoring devices, such as devices monitoring heart rate and blood pressure to portray better and quantify a patient's condition.

In conclusion, spoken and written speech represent distinct communication modalities, and accurate diagnosis and prognosis of speech, language, and communication disorders require an understanding of the unique characteristics of each. Continued research and collaboration between experts in AI, NLP, ML, acoustic analysis, and statistical modeling will further enhance our understanding and capabilities in assessing and treating speech, language, and communication disorders, ultimately improving the lives of individuals affected by these disorders. By considering these factors and leveraging technological advancements, clinicians and researchers can develop effective intervention plans and make informed prognostic judgments, ultimately improving the lives of individuals with speech, language, and communication disorders. The platform empowers clinicians to deliver effective and inclusive care to patients with speech, language, and communication impairments, ultimately improving their overall well-being.

**Tools Availability:** The tools are accessible online at the Open Brain AI's website: https://openbrainai.com.